\documentclass[runningheads]{llncs}
\usepackage{amsmath}
\usepackage{hyperref}
\usepackage{cleveref}
\usepackage{amssymb}
\usepackage{pifont}
\usepackage{array}
\usepackage{booktabs}
\usepackage{subcaption}
\usepackage[T1]{fontenc}
\usepackage{graphicx}
\begin{document}
\title{Data Augmentation via Latent Diffusion Models for Detecting Smell-Related Objects in Historical Artworks}
\titlerunning{Data Augmentation for Artwork Object Detection via LDMs}
\author{Ahmed Sheta \and
Mathias Zinnen \and
Aline Sindel \and
Andreas Maier \and
Vincent Christlein} 
\authorrunning{A. Sheta et al.}
%
\institute{Pattern Recognition Lab, Friedrich-Alexander Universität Erlangen-Nürnberg, Germany}
\maketitle              
\begin{abstract}
Finding smell references in historic artworks is a challenging problem. 
Beyond artwork-specific challenges such as stylistic variations, their recognition demands exceptionally detailed annotation classes, resulting in annotation sparsity and extreme class imbalance.
In this work, we explore the potential of synthetic data generation to alleviate these issues and enable accurate detection of smell-related objects.
We evaluate several diffusion-based augmentation strategies and demonstrate that incorporating synthetic data into model training can improve detection performance. 
Our findings suggest that leveraging the large-scale pretraining of diffusion models offers a promising approach for improving detection accuracy, particularly in niche applications where annotations are scarce and costly to obtain.  
Furthermore, the proposed approach proves to be effective even with relatively small amounts of data, and scaling it up provides high potential for further enhancements.
The source code for data generation and downstream evaluation is available at \url{https://github.com/ultiwinter/MT_DA_LDM_OD}.
\end{abstract}

\section{Introduction}
The sense of smell plays a crucial role in our everyday lives:
We are constantly surrounded by smells, mostly without noticing them. They help us remember, signal danger, and support communication. 
Despite its fundamental importance, smell has been overlooked in traditional art history and cultural heritage discourses~\cite{ehrich2021nose}. 
Recently, researchers across a wide range of disciplines have explored the cultural significance of smells~\cite{bembibre2022smelly}.
Specifically in art history, olfactory dimensions of artworks can not only reveal historical understandings of the sense, but also open up new interpretative dimensions of paintings~\cite{marx2023seeing}. 

Researchers from the Odeuropa project\footnote{\url{https://odeuropa.eu}} aim to uncover the olfactory dimensions of historic artworks through automatic extraction of visual smell references. 
However, recognizing such references is a complex task.
Recognition algorithms must address artwork-specific challenges, such as stylistic diversity, varying degrees of abstraction, and annotation sparsity.
Additionally, there are challenges specific to the domain of olfactory heritage. 
The diversity of smells often does not correlate with visual appearance, requiring algorithms to distinguish between fine-grained categories.
For example, two visually similar flowers might emit entirely different scents.
Moreover, smell is often not the focal point of a painting and is often only subtly and peripherally represented. 
As a result, smell-related objects tend to be small and spatially distributed across the entire canvas, contradicting the center-bias prevalent in common object detection benchmarks~\cite{zheng2024zone}.

To address these challenges, the Odeuropa researchers organized the ODeuropa Competition on Olfactory Object Recognition (ODOR)~\cite{odorchallenge} and introduced the similarly abbreviated Object Detection for Olfactory References dataset~\cite{zinnen2024smelly}.
While these efforts provide a benchmark for evaluating smell recognition algorithms and a foundation for model training, the fundamental problem of annotation sparsity remains.
The ODOR dataset contains only about 4,700 images, and although frequent classes such as "rose" are well represented, rare classes like "lobster" have fewer than 20 annotated instances. 

This work addresses the persisting challenges of data sparsity and class imbalance by synthesizing artificial training data. 
We evaluate several data generation strategies and demonstrate that even relatively small augmentations of the training set with synthetic data can improve detection performance. 
Our approach is scalable and can --with minor adaptations-- be applied to other applications where training data is limited and imbalanced.

\section{Related Work}
\paragraph{Computational Art Analysis \& Smell Reference Extraction}
The application of computer vision to the analysis of visual arts has a long-standing tradition, with applications across diverse areas such as art history~\cite{MADHU2023109153}, cultural heritage~\cite{willot2024creating}, search and retrieval~\cite{10.1093/llc/fqy006}, provenance research~\cite{lang_2025_14943104}, and image alignment~\cite{sindel2022artfacepoints}.
The results of automated art analysis can complement the traditional methodology in the humanities with a data-driven approach, enabling researchers to take a perspective of distant viewing~\cite{arnold2019distant}, either independently or in combination with multimodal data via knowledge graphs~\cite{10.1145/3594724}.

A major challenge for computer vision in artistic contexts is the representational gap between relatively uniform photographic object representations and the wide variety of artistic interpretations~\cite{hall2015cross}, further aggravated by varying levels of abstraction.
To bridge this domain gap and leverage pretraining from large-scale photographic datasets, researchers have employed domain adaptation techniques such as style transfer~\cite{MADHU2023109153,huang2024scene}, few-shot learning~\cite{9784141}, and weak supervision~\cite{gonthier2018weakly,mazzamuto2022weakly}.
Additionally, several artwork-focused approaches have been introduced for object detection~\cite{reshetnikov2022deart,zinnen2024smelly}, person detection~\cite{westlake2016detecting}, and human pose estimation~\cite{schneider2024poses,zinnen2024recognizing}.
However, these datasets are not comparable in scale to standard photographic datasets like COCO~\cite{lin2014microsoft}. 

For the specific task of detecting smell-related references, earlier work includes the recognition of smell-related objects~\cite{kim2018seeing}.
Within the Odeuropa project, the automatic recognition of smell-related gestures~\cite{zinnen2023sniffyart,hussian2024gesture,zinnen2024recognizing}, 
scenes~\cite{huang2024scene,10.1007/978-3-031-91572-7_10}, and the emotional context of olfactory imagery~\cite{patoliya2024smell} were explored.
Particular attention was paid to the detection of olfactory objects, with the ODOR challenge~\cite{odorchallenge} and dataset~\cite{zinnen2024smelly} providing benchmarks and training data tailored for detecting visual smell references.
Extracted visual references were then combined with textual references~\cite{menini2023scent} into a knowledge graph~\cite{lisena2022capturing} for further historical~\cite{leemans2022whiffstory} and museological~\cite{9e75c96be6ef4083a69a9a6bc65a2da5} interpretation.

Despite these efforts, existing approaches for visual smell reference extraction still suffer from general limitations in computational art analysis in general, including limited training data and the domain gap. 
This motivates our approach of generating synthetic data to overcome these obstacles.

\paragraph{Diffusion Models \& Data Augmentation}
Synthetic training data holds the potential to overcome both the scarcity of annotations and the representational domain gap by enabling the creation of virtually unlimited amounts of training images across diverse artistic styles. 
However, traditional generation methods such as GANs~\cite{goodfellow2014generative} often suffer from unstable training, poor image quality, and limited diversity~\cite{dhariwal2021diffusion}.
Diffusion models have addressed many of these issues.
Sohl-Dickstein et al.~\cite{sohl2015deep} introduced the idea of learning complex data distributions via reverse thermodynamic diffusion.
Building on this, Ho et al.~\cite{ho2020denoising} developed the Denoising Diffusion Probabilistic Model (DDPM), which enabled the generation of high-quality synthetic images.
Further improvements, such as optimized noise scheduling~\cite{nichol2021improved} and deterministic sampling~\cite{song2020denoising}, pushed the capabilities of these models, albeit with still high computational costs. 
Rombach et al.~\cite{rombach2022high} addressed this by shifting the diffusion process into latent space, encoding images with a variational autoencoder~\cite{kingma2013auto} and conditioning generation via CLIP~\cite{radford2021learning} embeddings. 
This allowed for precise control over outputs using textual prompts.
Extending this framework, Zhang et al.~\cite{zhang2023adding} introduced ControlNet, enabling spatial and structural guidance through depth maps, edge maps, segmentation masks, or pose annotations.

The controllability and quality of diffusion-generated images have prompted their application to data augmentation and training data generation.
Diffusion models have been used to include generated samples~\cite{toker2024satsynth}, interpolate between target classes~\cite{wang2024enhance}, manipulate high-level semantic attributes~\cite{trabucco2023effective}, generate visual priors~\cite{fang2024data}, and mix real and synthetic images~\cite{islam2024diffusemix}.
However, to the best of our knowledge, no prior work has specifically targeted fine-grained, imbalanced categories as required for smell-related object detection in artworks.


\section{Methodology}

\subsubsection{Preliminary Experiments}
\label{subsec:prelim_experiments}

In the initial phase of this study, we explored multiple diffusion-based data augmentation strategies listed in \cref{tab:inpainting_strategies}, each employing different masking strategies to determine the area for inpainting.
To assess their effectiveness, we conducted a preliminary experiment:
Using each strategy, we generated synthetic training sets of the same size as the original ODOR training set and evaluated the results both qualitatively and quantitatively.
Qualitative evaluation involved visually inspecting the coherence and plausibility of the generated images.
Quantitatively, we trained object detection models on each synthetic dataset and measured their performance on a fixed validation split from the real ODOR dataset. Detailed results are reported in \cref{sec:modelselectionresults}.

\begin{table}[t]
\centering
\caption{Overview of preliminary inpainting strategies and their shorthand labels. Methods marked with SD use Standard Stable Diffusion for inpainting, CTRL marks the usage of ControlNet~\cite{zhang2023adding}.}
\label{tab:inpainting_strategies}

\resizebox{\textwidth}{!}{%
\begin{tabular}{ll>{\raggedright\arraybackslash}p{10.8cm}}
\toprule
\textbf{Shorthand} & \textbf{Inpainting} & \textbf{Description} \\
\midrule
\textsc{ADAPT} & SD&  Adaptive entropy-based masking: masks the high-entropy half of the object bounding box to prioritize informative object regions while ensuring mask contiguity. \\
\textsc{ENT-H} & SD & High-entropy masking: masks high-information pixels based on local entropy values; often results in scattered, incoherent masks. \\
\textsc{ENT-L} & SD & Low-entropy masking: masks uniform, low-information regions; adding more diversity to these regions and complementing the high-information regions. \\
\textsc{SAL-H} & SD & High saliency masking: targets regions with strong gradient responses to enforce reconstruction of visually dominant object features. \\
\textsc{SAL-L} & SD & Low saliency masking: masks smoother areas with low gradient values to enhance diversity in non-salient regions. \\
\textsc{OPBG} & SD & Object-preserving background masking: randomly masks background patches while keeping all annotated objects intact to diversify contextual information. \\
\textsc{BORDER} & SD & Border region masking: masks the edge regions around objects to improve structural boundary learning. \\
\textsc{EDGE} & CTRL \& SD & Edge-controlled object generation combined with context generating class-balancing.\\
\bottomrule
\end{tabular}%
}
\end{table}

However, quantitative evaluation alone is insufficient for selecting the optimal generation strategy. 
Since initial experiments exclude original training data, methods that introduce fewer deviations from the original images might show artificially high performance. At the same time, those generating more variation could prove more effective when combined with real data or scaled to larger training sets.
To account for this bias, we adopted a heuristic selection approach that considers both the degree of deviation from the original data and the conceptual consistency of the generated images. 
Developing a more principled and less subjective selection methodology is left to future work.
Based on this heuristic, we selected four strategies for further exploration.
Of these, \textsc{OPBG}, \textsc{ADAPT}, and \textsc{ENT-L} failed to improve performance when scaled to larger synthetic datasets.
The remaining strategy, Edge-Conditioned Object Generation with Contextual Inpainting (\textsc{EDGE}), produced a measurable improvement and is described in detail below.

\subsubsection{Edge-Conditioned Object Generation with Contextual Inpainting}
Our synthetic data generation strategy is two-fold: 
First, to introduce additional stylistic diversity into the dataset, we replace all objects in the ODOR training set with synthetically generated counterparts. 
Second, to address class imbalance, we superimpose synthetic versions of underrepresented classes onto artificially generated backgrounds. 
\paragraph{Edge-Conditioned Object Replacement}

To augment the visual diversity of the dataset while preserving scene structure, we replace each object in the ODOR training set with a synthetically generated counterpart using ControlNet~\cite{zhang2023adding}. For each object, we extract its bounding box crop and generate an edge map using the Holistically-nested Edge Detection (HED) detector~\cite{xie2015holistically}. These edge maps are used as conditioning inputs to the ControlNet model, which guides a pretrained latent diffusion model to generate a structurally aligned but stylistically varied version of the object.

\begin{figure}[h]
    \centering
    \includegraphics[width=0.8\textwidth]{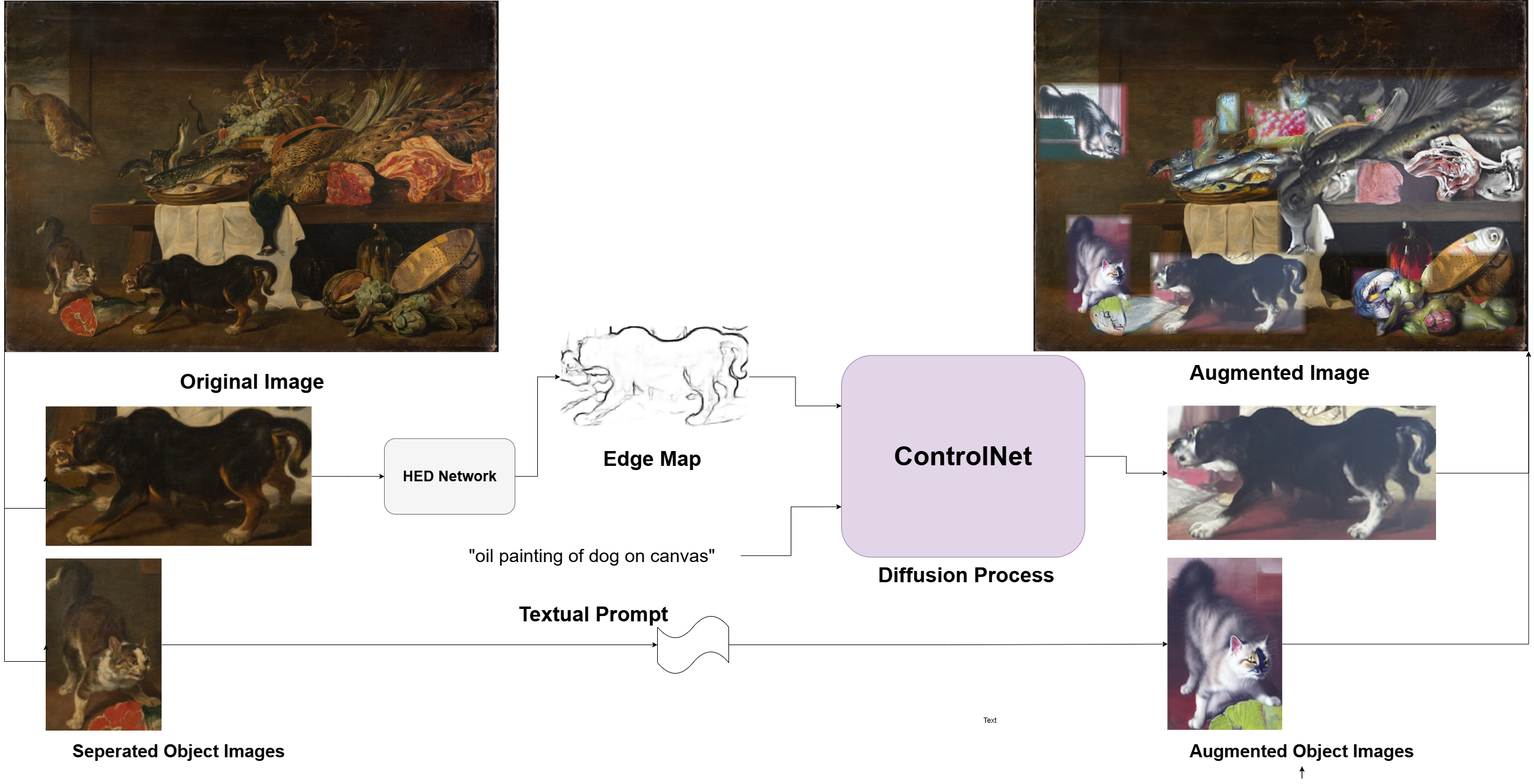}
    \caption{Illustration of the ControlNet-based augmentation pipeline. Each object is extracted from the ODOR images using its bounding box, processed through an HED edge detector to generate an edge map, and conditioned alongside a textual prompt in ControlNet. The augmented object is then reintegrated into its original position within the full ODOR image, preserving spatial and structural consistency.}
    \label{fig:controlnet_pipeline}
\end{figure}

The generation process is further conditioned using textual prompts of the form \textit{oil painting of \{class\_name\} on canvas} to maintain semantic consistency with the object category and align the visual output with the historical domain. 
To suppress undesirable features and enforce anatomical coherence, negative prompts (e.g., \textit{bad anatomy, bad structure}) are included. 
After generation, the synthetic object is blended back into its original position within the artwork image using boundary smoothing to mitigate sharp edges at the overlay boundary. An illustration of this pipeline is shown in Figure~\ref{fig:controlnet_pipeline}.

\paragraph{Class-Balancing} 

To address class imbalance, we selectively oversample underrepresented categories by generating ControlNet-based synthetic object crops, which are then placed onto blank canvases. 
Each underrepresented class is upsampled to ensure a minimum of 1,000 instances in the training set. This targeted augmentation strategy aims to diversify object representation without duplicating data or disrupting the domain's stylistic coherence. Per-class augmentation statistics are summarized in Table~\ref{tab:class_balancing_summary}.
\begin{table}[t] 
\centering
\caption{Tiered augmentation schedule to balance underrepresented classes toward 1000 instances.}
\scriptsize
\begin{tabular}{p{1.8cm}*{14}{>{\centering\arraybackslash}p{0.6cm}}}
\toprule
\textbf{Max. Inst.} & 5 & 9 & 19 & 29 & 49 & 74 & 99 & 149 & 249 & 499 & 999 & 3000 \\
\midrule
\textbf{Augs./Inst.} & 335 & 130 & 80 & 45 & 35 & 20 & 15 & 10 & 7 & 3 & 2 & 1 \\
\bottomrule
\end{tabular}
\label{tab:class_balancing_summary}
\end{table}

\begin{figure}[h]
    \centering
    \includegraphics[width=0.8\textwidth]{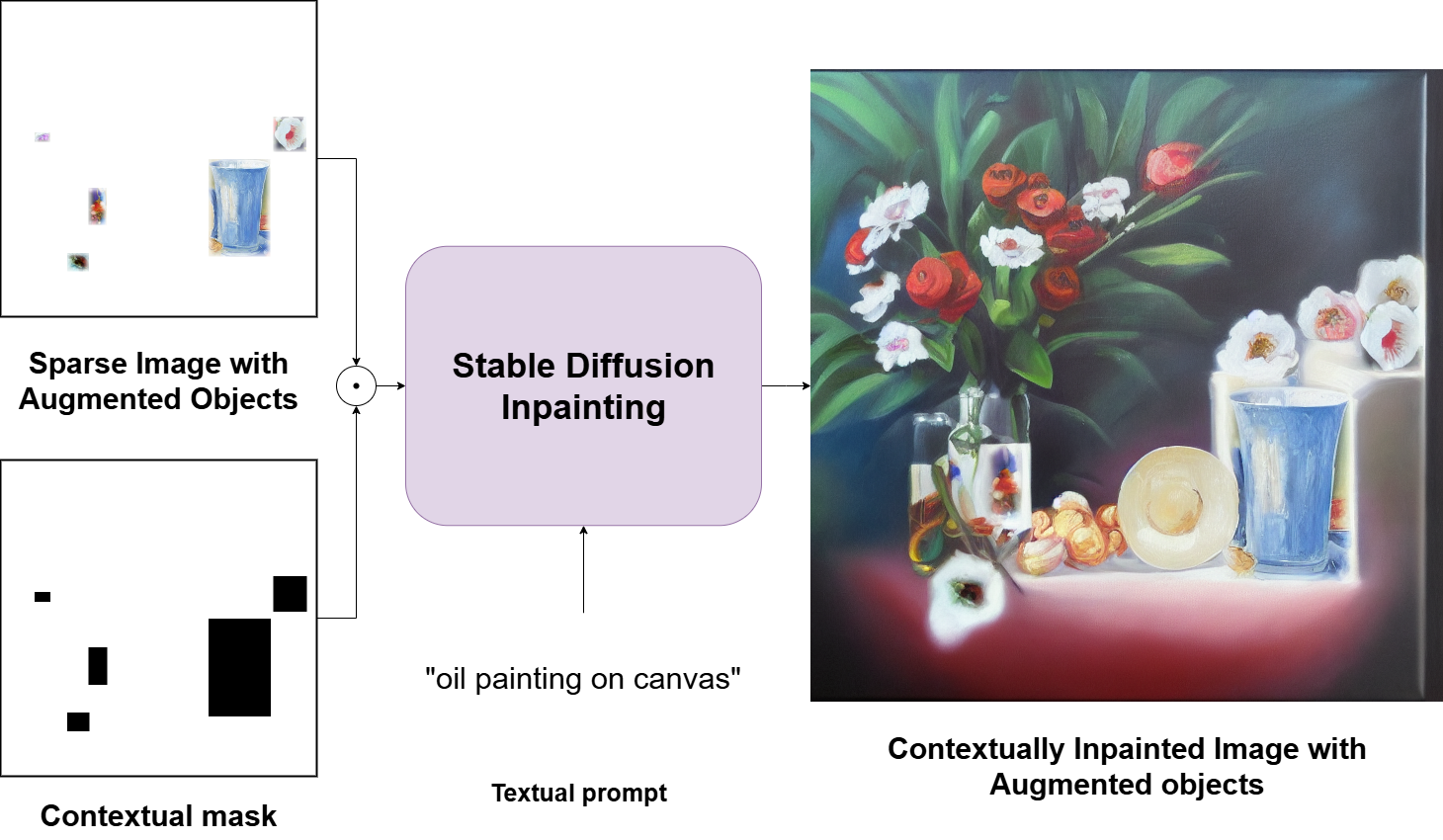}
    \caption{Illustration of the inpainting-based background generation for synthetic images. The process fills the blank background onto which ControlNet-augmented objects were previously overlaid, aiming to obtain realistic contextual surroundings. }
    \label{fig:blank_filling}
\end{figure}

\paragraph{Contextual Inpainting for Blank Placement}  
To prevent object detection models from having localization bias within the objects placed in blank images, ControlNet-generated object crops placed on blank images were post-processed using Stable Diffusion inpainting. As \cref{fig:blank_filling} shows, this step filled the surrounding canvas with contextually plausible background textures, ensuring the augmented samples remained visually coherent and semantically natural. Reducing the visual clutter on the objects' border would alleviate the localization bias that the object detection model might develop in case of sparse images.

\paragraph{ControlNet Finetuning}  
The pretrained ControlNet~\cite{zhang2023adding} was finetuned for 20 epochs on a curated set of ODOR object crops paired with edge maps (from the HED detector~\cite{xie2015holistically}) and category-specific prompts (\textit{oil painting of \{class\_name\} on canvas}) to adapt ControlNet to ODOR’s fine-grained and historically specific object classes. 

This pipeline, combining edge-controlled fine-tuned ControlNet generation with structured integration and contextual inpainting, or short \textsc{EDGE}, provided a scalable and controllable mechanism for dataset expansion, particularly effective for improving the representation of rare and complex object classes in artistic imagery. 

The computational cost of ControlNet-based data Augmentation amounted to 384 GPU hours using NVIDIA Tesla V100 GPUs. 
This includes 346 GPU hours for generating object-centric image samples, conducted in parallel across 4 GPUs, each running for approximately 86.5 hours. 
An additional 38 GPU hours were spent on contextual inpainting using Stable Diffusion v1.5.

\section{Results}
\label{sec:results}

\subsubsection{Experimental Setup}

To evaluate the downstream detection task, we adopt Ultralytics' YOLOv11-M architecture~\cite{khanam2024yolov11}. 
All experiments follow the official default configuration except for the learning rate, which was set to 0.001 during, and reduced to 0.0007 in cases of second-stage finetuning to mitigate the catastrophic forgetting.
To evaluate the performance of the different configurations, we use the original test split provided by~\cite{zinnen2024smelly}. 
For hyperparameter tuning, model selection, and preliminary evaluations, we further split a validation set from the ODOR training set, resulting in 3408 images for training, 856 for validation, and 448 for testing.
All evaluations are reported using the COCO-style mean Average Precision (mAP@0.5:0.95) as defined in~\cite{lin2014microsoft}.
To mitigate the effect of randomness during training, we report all performance metrics as the mean and standard deviation over three independent training runs.

\subsubsection{Baseline Comparison}

\begin{table}[t]
\centering
\scriptsize
\caption{Results comparing the baseline with selected augmentation strategies. Values are mAP@0.5:0.95 percentages reported as mean ± standard deviation over three independent runs. The percentage change relative to the baseline is reported in brackets.}
\label{tab:final_augmentation_results}
\begin{tabular}{p{3.3cm} c r r}
\toprule
\textbf{Method} & \textbf{Real : Synthetic (\%)} & \textbf{Val mAP} & \textbf{Test mAP}  \\
\midrule
Baseline (Vanilla) & $100\phantom{.} : \phantom{0}0.0$ & $17.7 \pm 0.2$ & $16.3 \pm 0.5$ \\
\midrule
\textbf{EDGE} & $14.1 : 85.9$ & $\mathbf{18.4 \pm 0.1\ (+4.0\%)}$ & $\mathbf{17.4 \pm 0.2\ (+6.7\%)}$  \\
OPBG & $50.0 : 50.0$ & $17.0 \pm 0.1$ (\phantom{a}-4.0\%) & $15.9 \pm 0.2$ (\phantom{a}-2.5\%)  \\
ADAPT & $37.4 : 62.6$ & $15.8 \pm 0.1$ (-10.8\%) & $14.5 \pm 0.2$ (-11.1\%) \\
ENT-L & $37.4 : 62.6$ & $15.2 \pm 0.1$ (-14.2\%) & $13.9 \pm 0.4$ (-14.3\%) \\
\bottomrule
\end{tabular}
\end{table}

Table~\ref{tab:final_augmentation_results} summarizes the final detection performance of YOLOv11-M models trained with different augmentation strategies. 

The baseline model, trained solely on the original ODOR dataset, hereafter referred to as \textsc{Vanilla}, achieved a test mAP of 16.3\%. Among the augmentation methods, the \textsc{EDGE} approach yielded the best performance, improving test accuracy by 6.7\% relative to the \textsc{Vanilla} baseline. 

\begin{figure}
\begin{subfigure}[t]{.5\textwidth}
    \centering
    \includegraphics[width=\linewidth,height=4cm]{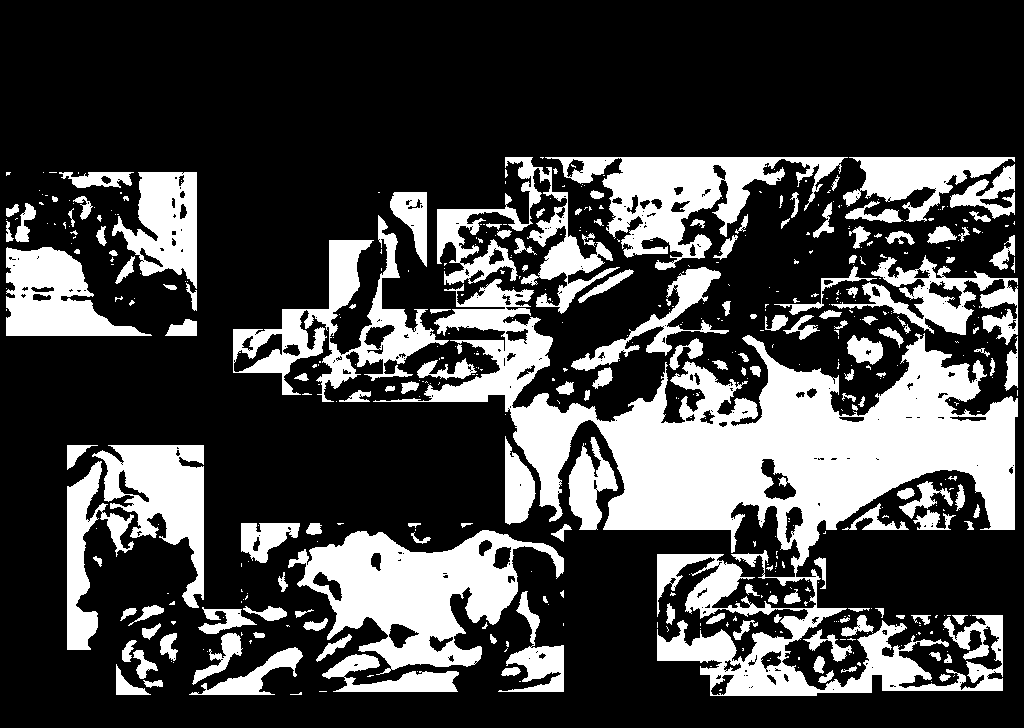}
    \caption{Generated Mask}
\end{subfigure}
\begin{subfigure}[t]{.5\textwidth}
    \centering
    \includegraphics[width=\linewidth,height=4cm]{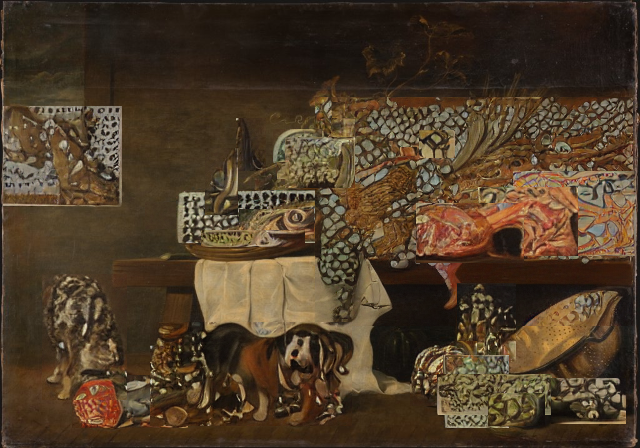}
    \caption{Inpainted Image}
\end{subfigure}
    \caption{Example image, generated by the low entropy masking strategy (ENT-L).}
    \label{fig:fragmented}
\end{figure}

In contrast, inpainting-based methods showed limited effectiveness. Augmenting only the background (\textsc{OPBG}) or using entropy-guided object masking (\textsc{ADAPT}) resulted in negative gains. The poorest performance was observed with \textsc{ENT-L}, which masks low-entropy regions; this likely stems from the fragmented and non-contiguous nature of the generated masks, which the inpainting model fails to handle coherently as illustrated in \cref{fig:fragmented}. 
These results suggest that diffusion-based inpainting models are not well-suited for reconstructing fine-grained pixel fragments. Instead, they require spatially contiguous masks to produce semantically meaningful outputs.
To address this issue by enforcing block-like, high-entropy masking, we introduced the \textsc{ADAPT} strategy. However, its downstream performance remained suboptimal.

\subsubsection{Additional Analyses}

\paragraph{Training Scheme Comparison}

To evaluate the impact of training order, we compared three finetuning schemes: (1) joint training on original and augmented data (mixed), (2) finetuning first on augmented data and then on original data (aug$\rightarrow$orig), and (3) the reverse—finetuning on original data followed by augmented data (orig$\rightarrow$aug). 

The results for the mixed training strategy are reported in Table~\ref{tab:final_augmentation_results}. 
Table~\ref{tab:TrainAugftOrig_results} presents results for the second scheme (aug$\rightarrow$orig). Only the \textsc{EDGE} strategy improved over the baseline, achieving a test mAP of 16.9\%. 
These results highlight the polarizing nature of joint training, where performance depends heavily on the semantic and structural quality of the augmentations. The initial synthetic-only stage also converged early, with training typically stopping between epochs 4 and 9.

\begin{table}[t]
\centering
\scriptsize
\caption{Results of training on augmented images (including class balancing) and finetuning on original images. Metrics reported are mean ± standard deviation over 3 runs.}
\begin{tabular}{l@{\hspace{1cm}}r@{\hspace{1cm}}r}
\toprule
\textbf{Method} & \textbf{Val mAP} & \textbf{Test mAP} \\
\midrule
\multicolumn{3}{c}{\textit{1. Stage: Finetuning on Augmented Images of}} \\
EDGE & $4.1 \pm 0.2$ & $4.1 \pm 0.3$ \\
ADAPT & $3.6 \pm 0.1$ & $4.0 \pm 0.3$ \\
ENT-L & $3.4 \pm 0.1$ & $03.5 \pm 0.4$ \\
\textbf{OPBG} & \textbf{$14.8 \pm 0.1$} & \textbf{$14.0 \pm 0.3$} \\
\midrule
\multicolumn{3}{c}{\textit{2. Stage: Finetuning on Original Images}} \\
\textbf{EDGE} & \textbf{$18.5 \pm 0.6$ (+4.6\%)} & \textbf{$16.9 \pm 0.3$ (+3.6\%)} \\
ADAPT & $17.9 \pm 0.6$ (\phantom{0}+1.1\%) & $16.1 \pm 0.3$ (\phantom{0}–1.2\%) \\
ENT-L & $17.1 \pm 0.6$ (\phantom{0}–3.4\%) & $15.2 \pm 0.6$ (\phantom{0}–3.6\%) \\
OPBG  & $16.4 \pm 0.6$ (\phantom{0}–7.3\%) & $15.4 \pm 0.3$ (\phantom{0}–6.7\%) \\
\bottomrule
\end{tabular}
\label{tab:TrainAugftOrig_results}
\end{table}

In contrast, the third scheme (orig$\rightarrow$aug) proved ineffective across all methods. In these cases, validation performance consistently declined during the second stage, and early stopping reverted to the best model checkpoint saved at epoch 1, indicating that continued training on synthetic data led to overfitting or distribution shift. These findings reinforce that synthetic data is most beneficial when used concurrently with real data, rather than in staged isolation.

\paragraph{Preliminary Experiment Results}
\label{sec:modelselectionresults}

To identify the most promising augmentation strategies, we conducted preliminary experiments by training YOLOv11-M solely on synthetic data (without class balancing) and evaluating on the original ODOR validation and test sets. As shown in Table~\ref{tab:prelim_augmentation_results}, \textit{EDGE}, \textit{ENT-L}, and \textit{ADAPT} emerged as the top-performing object-aware strategies, while \textit{OPBG} outperformed all others in the context-aware category. These four methods were selected for subsequent large-scale experiments.

\begin{table}[t]
\centering
\caption{Preliminary results of training YOLO-M on the augmented ODOR images only from every augmentation strategy, without class balancing, evaluated on the original validation and test sets. Values are reported as mean ± standard deviation over 3 runs.}
\scriptsize
\label{tab:prelim_augmentation_results}
\begin{tabular}{l@{\hspace{1cm}}r@{\hspace{1cm}}r}
\toprule
\textbf{Method} & \textbf{Val mAP} & \textbf{Test mAP} \\
\midrule
\textbf{EDGE}            & $\mathbf{2.9 \pm 0.4}$ & $\mathbf{3.0 \pm 0.3}$ \\
\textbf{ENT-L}      & $\mathbf{2.5 \pm 0.2}$ & $\mathbf{2.6 \pm 0.2}$ \\
\textbf{ADAPT}      & $\mathbf{2.1 \pm 0.2}$ & $\mathbf{2.4 \pm 0.3}$ \\
SAL-H              & $1.8 \pm 0.3$ & $2.2 \pm 0.2$ \\
ENT-H               & $1.4 \pm 0.2$ & $1.2 \pm 0.1$ \\
SAL-L            & $1.3 \pm 0.2$ & $1.1 \pm 0.2$ \\
\midrule
\textbf{OPBG}     & $\mathbf{14.8 \pm 0.1}$ & $\mathbf{14.0 \pm 0.3}$ \\
BORDER         & $12.8 \pm 0.3$ & $12.2 \pm 0.1$ \\

\bottomrule
\end{tabular}
\end{table}

\section{Conclusion}

Identifying smell references in historic artworks remains a challenging task due to the combination of stylistic variation, fine-grained semantic classes, and annotation sparsity.

In this work, we explored diffusion-based data augmentation methods to address these limitations.
We systematically evaluated a range of masking strategies for inpainting and identified one particularly promising approach: edge-based conditioning combined with contextual inpainting (\textsc{EDGE}). 
This method demonstrated measurable improvements in detection performance, even when applied to relatively small synthetic training sets.

While the observed performance gains are limited, they suggest the potential of this technique, particularly when scaled to generate a larger number of synthetic images.
Given its generality, we anticipate that this method to be applicable in other domains where annotated data is scarce or imbalanced. 
Scaling up the synthetic data generation and applying the approach to other domains are natural and promising extensions to the current work.

Further improvements may be achieved by refining the masking strategies used during inpainting and developing adaptive training schemes. 
In particular, gradually increasing the proportion of real data relative to synthetic data over the training schedule may allow models to better adapt to the real data distribution while retaining the increased variability introduced by synthetic data.

Overall, our findings highlight the potential of diffusion-based augmentation to enrich niche datasets and offer a step forward in the computational analysis of olfactory references in historical art, which aligns well with recent efforts to uncover olfactory aspects of cultural heritage.

%
%
%
\bibliographystyle{splncs04}
\bibliography{mymain}

\end{document}